\documentclass{article} % For LaTeX2e
\usepackage[OT1]{fontenc}
\usepackage{iclr2027_conference,times}

\usepackage{amsmath,amsfonts,bm}

\def\eqref#1{equation~\ref{#1}}
\def\1{\bm{1}}

\DeclareMathAlphabet{\mathsfit}{\encodingdefault}{\sfdefault}{m}{sl}
\SetMathAlphabet{\mathsfit}{bold}{\encodingdefault}{\sfdefault}{bx}{n}

\usepackage{amsmath,amssymb,mathtools,amsthm}
\usepackage{booktabs,multirow,array,tabularx}
\usepackage{graphicx,xcolor}
\usepackage{microtype}
\usepackage{algorithm}
\usepackage{algorithmic}
\usepackage{hyperref}

\newcommand{\method}{Waypoint-OPD}
\newcommand{\directopd}{Direct-OPD}
\newcommand{\opdtwo}{\textsc{OPD$^2$}}
\newcommand{\student}{\pi_{\theta}}
\newcommand{\studentref}{\pi_{S}}
\newcommand{\teacherpre}{\pi_{0}}
\newcommand{\teachermid}{\pi_{m}}
\newcommand{\teacherpost}{\pi_{1}}
\newcommand{\cK}{\mathcal{K}}
\newcommand{\sg}{\operatorname{sg}}

\title{\method: Transferring RL-Induced Policy Shifts via Gated On-Policy Distillation}

\author{Bingxi Hou, 
Guochao Jiang, 
Guofeng Quan, 
Weiqing Li, 
Wenfeng Feng, \\
\textbf{Guohua Liu}, 
\textbf{Yuewei Zhang}\footnotemark[1]\thanks{Corresponding author.} \\
Alibaba Cloud Computing \\ \texttt{\{anyue.jgc, liyou.zyw\}@alibaba-inc.com}
}

\begin{document}
\iclrfinalcopy
\maketitle

\begin{abstract}
Policy-shift distillation reuses the difference between a teacher's log probabilities before and after reinforcement learning (RL) as supervision on student-generated trajectories. However, this endpoint policy shift cannot reveal whether local
action preferences evolved consistently during teacher training. The same endpoint shift can result from early and late shifts that reinforce or partially cancel each other. We investigate whether agreement between early and late teacher policy shifts can guide selective transfer. We introduce \textbf{Waypoint-guided On-Policy Distillation}(\method), which uses an intermediate teacher checkpoint from the same RL run to compare early and late shifts in relative action preferences at student-visited prefixes. We gate the endpoint shift according to the directional agreement between the early and late shifts, suppressing it when they point in opposing directions. Compared with endpoint-only transfer, our method requires only one additional frozen-teacher scoring pass. Using RL-trained Qwen3-4B teachers, we evaluate our method with Qwen3-8B, Qwen3-14B, and Qwen3-30B-A3B as student models in mathematical and code reasoning. In all six domain--student settings, \method{} achieves the highest domain-average score among the compared distillation methods. Ablations with the Qwen3-8B student show that removing or shuffling gates reduces mathematical reasoning performance. These results support the use of directional agreement between early and late teacher policy shifts to weight the endpoint shift during student training. Our code is available at 
\url{https://github.com/hhhbxloveai/WaypointOPD}.
\end{abstract}

\begin{figure}[!h]
  \centering
  \includegraphics[width=\linewidth]{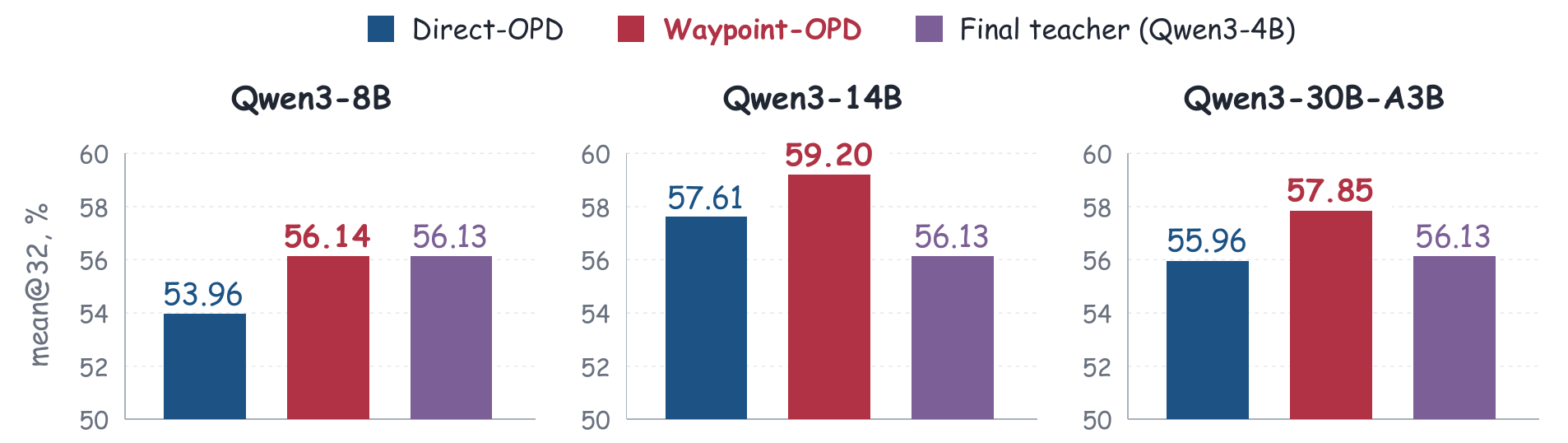}
  \caption{\textbf{Mathematical reasoning across three student models.} \texttt{mean@32} accuracy (\%), averaged over seven benchmarks. \method{} improves over \directopd{} by 2.18, 1.59, and 1.89 percentage points, respectively. Teacher scores are from the Final Qwen3-4B checkpoint.}
  \label{fig:overview}
\end{figure}

\section{Introduction}
\label{sec:introduction}

Reinforcement learning can improve language-model instruction following using rewards learned from human preferences \citep{ouyang2022instructgpt}, and mathematical and code reasoning using verifiable rewards \citep{shao2024deepseekmath,guo2025deepseekr1,yu2025dapo}. The latter requires repeated generation, verification, and optimization of on-policy responses. Reusing a teacher's RL improvements offers a way to share this training effort across target models. This raises the question: which local changes in the teacher's policy should guide a student on its own generated
trajectories?

Standard on-policy distillation (OPD) trains the student to match the teacher's token distributions on student-generated prefixes \citep{agarwal2024gkd}. Its supervision remains tied to the teacher's absolute policy and capability limitations.
On-Policy Delta Distillation (\opdtwo{}) uses the change between a reasoning-tuned teacher and its base model, retaining token-level signals when the centered change agrees in sign with the centered standard distillation signal \citep{heo2026opd2}. Direct On-Policy Distillation (\directopd{}) focuses on RL-induced improvements. It uses the log-probability ratio between pre-RL and post-RL teacher checkpoints as a dense implicit reward on student-generated states, transferring the teacher's policy change without matching its absolute distribution
\citep{feng2026directopd}. Both methods extract changes from a pair of teacher models, but neither directly measures whether consecutive changes during teacher RL point in compatible directions.

The Initial-to-Final log-probability ratio captures only the change between the teacher's policies before and after RL training. The same endpoint signal can arise from early and late policy changes that point in similar or opposing directions. Endpoint-only transfer cannot distinguish these cases. We hypothesize that agreement between the early and late changes provides evidence for deciding how much weight to assign an endpoint shift at a student-generated prefix.

Building on this hypothesis, we propose \method{}, which uses agreement between consecutive teacher shifts to gate the endpoint transfer signal. We retain three checkpoints from the same teacher RL run: Initial before RL, Waypoint at an intermediate RL step, and Final at the end of RL. At each student-generated prefix, we select its top-$k$ most probable next-token candidates. Over these candidates, we compare how the teacher's relative preferences change from Initial to Waypoint and from Waypoint to Final. Their agreement gates the centered Initial-to-Final signal, retaining aligned changes and suppressing reversals. Waypoint thus supplies evidence for weighting the endpoint shift, which remains the transferred signal. Relative to endpoint-only transfer, this requires one additional frozen teacher scoring pass, with no new teacher training or verifier calls during student transfer.

Our contributions are:
\begin{itemize}
  \item \textbf{Temporal agreement as a criterion for selective transfer.} We use the Waypoint checkpoint to assess temporal agreement at student-generated prefixes, revealing directional information unavailable from the endpoint pair alone.
  \item \textbf{Waypoint-guided gating of endpoint signals.}
  We develop \method{}, a dense on-policy objective that uses this agreement to weight the endpoint signal at each student-generated prefix, with one additional frozen teacher scoring pass.
  \item \textbf{Higher average performance across domains and model sizes.} We evaluate Qwen3-8B, Qwen3-14B, and Qwen3-30B-A3B\citep{yang2025qwen3} on mathematics and code. \method{} achieves the highest recorded domain average among the evaluated baselines in all six settings
  (Figure~\ref{fig:overview} and Tables~\ref{tab:full-math-results} and~\ref{tab:code-summary}). Ablations on Qwen3-8B mathematical reasoning support the value of temporal gating
  (Table~\ref{tab:gate-ablation}).
\end{itemize}

\section{Preliminaries}
\label{sec:preliminaries}

\subsection{Problem setup and notation}
\label{sec:setup}

Let $x\sim\mathcal{D}$ be a prompt, $y=(y_1,\ldots,y_T)$ a response, and $s_t=(x,y_{<t})$ the prefix at which a candidate token $a$ is scored. The student $\student$ generates on-policy responses $y\sim\student(\cdot\mid x)$. Its initialization $\studentref$ remains fixed as a reference during transfer.

We denote the policies of the Initial, Waypoint, and Final teachers by $\teacherpre$, $\teachermid$, and $\teacherpost$, respectively. All three checkpoints come from the same teacher RL run: Initial before RL, Waypoint at an intermediate step, and Final after RL. They remain frozen during transfer. We measure the teacher's RL-induced change relative to $\teacherpre$ and regularize student updates relative to student initialization $\studentref$. Teachers and student share a tokenizer, allowing direct comparison of probabilities for the same token.

\subsection{On-Policy Distillation}
\label{sec:on-policy-supervision}

On-policy distillation (OPD) samples responses from the student and aligns its next-token distribution with the teacher's at the prefixes the student visits \citep{agarwal2024gkd}. Using the Final teacher $\teacherpost$, OPD minimizes the reverse-KL objective
\begin{equation}
  \mathcal{L}_{\text{OPD}}(\theta)
  =\mathbb{E}_{x\sim\mathcal{D},\,y\sim\student(\cdot\mid x)}
  \left[\sum_{t=1}^{T}
  D_{\text{KL}}\!\left(
    \student(\cdot\mid s_t)\,\Vert\,\teacherpost(\cdot\mid s_t)
  \right)\right].
  \label{eq:opd-objective}
\end{equation}

\subsection{Direct On-Policy Distillation}
\label{sec:endpoint-transfer}

\directopd{} uses the teacher's RL-induced policy shift as an implicit reward on student-generated responses \citep{feng2026directopd}. The Initial-to-Final shift decomposes into token-level log-probability ratios:
\begin{equation}
  \Delta^{(0,1)}(y\mid x)
  =\log\frac{\teacherpost(y\mid x)}{\teacherpre(y\mid x)}
  =\sum_{t=1}^{T}\delta_t(y_t),
  \label{eq:endpoint-shift}
\end{equation}
where
\begin{equation}
  \delta_t(a)
  =\log\teacherpost(a\mid s_t)-\log\teacherpre(a\mid s_t).
  \label{eq:endpoint-token-shift}
\end{equation}
A positive shift indicates that teacher RL increased a token's probability at $s_t$; a negative shift indicates a decrease. At an idealized teacher optimum with KL regularization toward $\teacherpre$, the sequence log-ratio recovers the reward
up to positive scale and a prompt-dependent constant \citep{rafailov2023dpo} as used by \citet{feng2026directopd}. This motivates the endpoint signal; we do not assume that finite-step teacher RL attains this optimum.

The idealized sequence-level objective maximizes this shift while
regularizing the student toward its own initialization:
\begin{equation}
  \mathcal{J}_{\text{Direct-OPD}}(\theta)
  =\mathbb{E}_{x\sim\mathcal{D}}\!\left[
    \mathbb{E}_{y\sim\student(\cdot\mid x)}
      [\Delta^{(0,1)}(y\mid x)]
    -\beta D_{\text{KL}}\!\left(
      \student(\cdot\mid x)\,\Vert\,\studentref(\cdot\mid x)
    \right)\right],
  \label{eq:direct-opd-objective}
\end{equation}
where $\beta>0$ controls the student KL penalty. In practice, \directopd{} uses $\delta_t(a)$ as an immediate token-level reward at student-generated prefixes, without verifier queries.

\section{Method}
\label{sec:method}

\begin{figure}[!htbp]
  \centering
  \includegraphics[width=\linewidth]{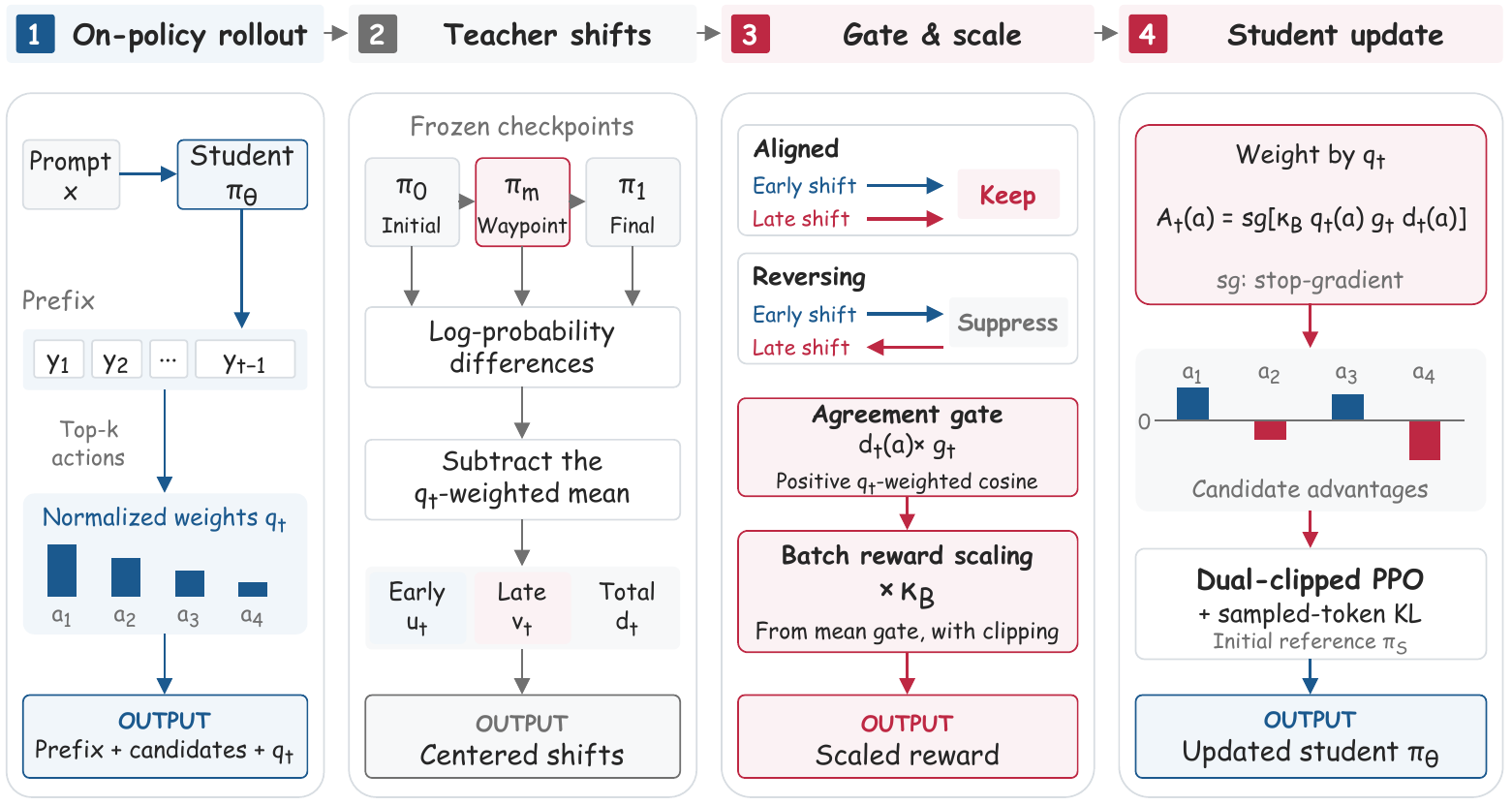}
  \caption{\textbf{Training pipeline of \method{}.}
  Initial, Waypoint, and Final score candidates at student-generated prefixes. Their centered segment shifts define an agreement gate on the endpoint shift. Batch scaling and detached candidate advantages drive the student update Waypoint measures consistency only.}
  \label{fig:waypoint-pipeline}
\end{figure}

\method{} augments the endpoint signal in Section~\ref{sec:endpoint-transfer} with the Waypoint checkpoint $\teachermid$ from the same teacher RL run. Agreement between the early and late shifts weights the endpoint signal on
student-visited prefixes. Waypoint supplies consistency evidence rather than an imitation target. Figure~\ref{fig:waypoint-pipeline} shows the pipeline.

\subsection{Student-local trajectory consistency}
\label{sec:trajectory-consistency}

At a student-visited prefix $s_t$,
$\cK_t=\operatorname{TopK}(\student(\cdot\mid s_t),k)$ selects the $k$ token candidates with the highest student probabilities. The normalized student distribution on this support is
\begin{equation}
  q_t(a)=
  \frac{\student(a\mid s_t)}
       {\sum_{b\in\cK_t}\student(b\mid s_t)},
  \qquad a\in\cK_t.
  \label{eq:student-topk}
\end{equation}
These weights focus the comparison on likely student actions.
Consistency is evaluated on the local support that receives the student update, with more probable candidates receiving greater
weight. We score these candidates with all three frozen teachers.
The two consecutive teacher shifts are
\begin{align}
  \delta^{(1)}_t(a)
    &=\log\teachermid(a\mid s_t)-\log\teacherpre(a\mid s_t), \\
  \delta^{(2)}_t(a)
    &=\log\teacherpost(a\mid s_t)-\log\teachermid(a\mid s_t),
  \label{eq:segment-shifts}
\end{align}
and telescope to $\delta_t=\delta^{(1)}_t+\delta^{(2)}_t$.

A uniform shift across candidates leaves their relative preferences unchanged, yet can affect the angle between raw shift vectors. We remove this action-independent component with the $q_t$-centering operator
\begin{equation}
  C_{q_t}[z](a)=z(a)-\mathbb{E}_{b\sim q_t}[z(b)]
  \label{eq:q-centering}
\end{equation}
and write $u_t=C_{q_t}[\delta^{(1)}_t]$,
$v_t=C_{q_t}[\delta^{(2)}_t]$, and
$d_t=C_{q_t}[\delta_t]=u_t+v_t$.
We weight candidates by student probability using the inner product
$\langle u_t,v_t\rangle_{q_t}=\sum_{a\in\cK_t}q_t(a)u_t(a)v_t(a)$.
The local temporal agreement and weighted norms are
\begin{equation}
  c_t=
  \frac{\langle u_t,v_t\rangle_{q_t}}
       {\lVert u_t\rVert_{q_t}\lVert v_t\rVert_{q_t}},
  \quad
  \lVert u_t\rVert_{q_t}=\sqrt{\langle u_t,u_t\rangle_{q_t}},
  \quad
  \lVert v_t\rVert_{q_t}=\sqrt{\langle v_t,v_t\rangle_{q_t}}.
  \label{eq:weighted-cosine}
\end{equation}

\subsection{Agreement gating and batch-scale normalization}
\label{sec:power-normalization}

When both centered segment norms exceed $\varepsilon=10^{-8}$, we use the positive-cosine gate
\begin{equation}
  g_t = \max(0,c_t).
  \label{eq:power-gate}
\end{equation}
If either norm is at most $\varepsilon$, we use the low-norm fallback $g_t=1$. The gate $g_t\in[0,1]$ is shared by all candidates at prefix $s_t$. Outside this fallback, it removes the endpoint signal when $c_t\leq0$ and scales its magnitude by $c_t$ otherwise. Algorithm and implementation conventions are given in Appendix~\ref{app:implementation}.

For illustration, consider two decompositions of a nonzero $q_t$-centered endpoint $d$, both outside the low-norm fallback. The pair $u=v=d/2$ gives $c=1$ and $g=1$, whereas $u=2d$, $v=-d$ gives $c=-1$ and $g=0$. Both yield $u+v=d$, so endpoint-only transfer cannot distinguish them.
Appendix~\ref{app:endpoint-preservation} derives the endpoint's
independence from Waypoint at fixed student prefixes and weights. Directional agreement alone does not guarantee beneficial supervision.

To stabilize the average gate scale across batches, we multiply all gated rewards in batch $B$ by
\begin{equation}
  \kappa_B =
  \operatorname{clip}\!\left(
    \frac{\tau}{\max\!\left(|B_{\text{pref}}|^{-1}
      \sum_{t\in B_{\text{pref}}}g_t,\varepsilon_{\text{num}}\right)},
    \kappa_{\text{min}},\kappa_{\text{max}}
  \right),
  \label{eq:gate-normalization}
\end{equation}
where $B_{\text{pref}}$ contains valid response prefixes in $B$, and $\varepsilon_{\text{num}}>0$ is a numerical floor preventing division by zero. The ratio targets a mean rescaled gate of $\tau>0$, while clipping restricts $\kappa_B$ to $[\kappa_{\text{min}},\kappa_{\text{max}}]$, with
$0<\kappa_{\text{min}}\leq\kappa_{\text{max}}$. This shared positive multiplier preserves relative gate weights and zero gates. The effective multiplier $\kappa_B g_t$ can exceed one; batch scaling does not fix the reward-vector norm.

\subsection{Dense candidate objective}
\label{sec:objective}

The final coefficient for candidate action $a\in\cK_t$ is
\begin{equation}
  A_t(a)=\sg\!\left[\kappa_B\,q_t(a)\,g_t\,d_t(a)\right].
  \label{eq:advantage}
\end{equation}
Following the signal-minus-KL form of Eq.~\ref{eq:direct-opd-objective}, the local on-policy surrogate to maximize is
\begin{equation}
  \begin{aligned}
  \mathcal{J}_{\text{WP}}(\theta)
  =\mathbb{E}_{x\sim\mathcal{D}}\!\Bigg[
    &\mathbb{E}_{y\sim\student(\cdot\mid x)}
      \left[\sum_{t=1}^{T}\sum_{a\in\cK_t}
        A_t(a)\log\student(a\mid s_t)\right] \\
    &-\beta\,\widehat{D}_{\text{KL}}\!\left(
      \student(\cdot\mid x)\,\Vert\,\studentref(\cdot\mid x)
    \right)\Bigg],
  \end{aligned}
  \label{eq:waypoint-objective}
\end{equation}
Rollout samples and candidate supports are held fixed when differentiating this surrogate, and the stop-gradient operator $\sg$ detaches all factors in $A_t$. Candidate credit is dense on $\cK_t$, while the KL penalty is estimated on sampled response tokens and anchors the student to $\studentref$.
The implementation uses dual-clipped PPO, which has the same gradient as this surrogate at the rollout policy (Appendix~\ref{app:implementation}).

Relative to \directopd{}, \method{} adds one frozen teacher
checkpoint and scoring pass, without additional trainable parameters or student rollouts. Derivations and signal properties appear in Appendix~\ref{app:proofs}.

\section{Experiments}
\label{sec:experiments}

We examine three questions:

\noindent\textbf{Q1. Does \method{} improve transfer across math, code, and model scales?}\par
\noindent\textbf{Q2. Does temporal consistency improve on endpoint transfer and Final-teacher imitation?}\par
\noindent\textbf{Q3. How do gate assignment, mean normalization, and Waypoint position affect performance?}\par

Supplementary training-time and response-length analyses appear in Appendix~\ref{app:efficiency}.

\subsection{Models, data, and evaluation}
\label{sec:experimental-setup}

\paragraph{Teacher checkpoints and student models.}
For each domain, we use three checkpoints from the same Qwen3-4B
\citep{yang2025qwen3} RL run: Initial ($\teacherpre$) before RL,
Waypoint ($\teachermid$) at RL step 600, and Final ($\teacherpost$) at RL step 1200. Math and code teachers use non-thinking RL. All three checkpoints remain frozen during transfer.
We train Qwen3-8B, Qwen3-14B, and Qwen3-30B-A3B students using the teacher's RL-induced policy shifts. These students span three scales and dense and MoE architectures within one family with compatible tokenizers. The transfer objective does not require the teacher to outperform the student. We report all three teacher checkpoints and the Base students in the result tables to show pre-transfer performance.

\paragraph{Training and evaluation data.}
Math transfer uses DAPO-Math-17K prompts \citep{yu2025dapo}. We evaluate on seven benchmarks: AIME 2024, AIME 2025, AIME 2026, HMMT February 2025, HMMT November 2025, HMMT February 2026, and AMC 2023. For each problem, \texttt{mean@32} measures
average correctness over 32 generated responses. Benchmark scores average these values over problems, and Avg. is the unweighted mean across benchmarks.

Code transfer uses a Codeforces-derived prompt set. Evaluation uses the 378-problem MBPP+ set from EvalPlus \citep{liu2023evalplus}, reporting both the original MBPP tests \citep{austin2021mbpp} and the original plus additional tests (MBPP+), together with LiveCodeBench v4, v5, and v6
\citep{jain2025livecodebench}.
We report \texttt{mean@8}, which averages correctness over eight generated responses per problem and then over the problems in each benchmark. Code Avg. is the unweighted mean over these five evaluation scores. Within each benchmark, methods share prompt templates, decoding parameters, maximum generation lengths, and sample counts.

\paragraph{Student checkpoint selection.}
For both domains, we use AIME 2024 as the validation set and select each trained student's checkpoint with the highest validation accuracy. All benchmark results for that student use this selected checkpoint.

\paragraph{Implementation.}
We build on \texttt{verl} \citep{sheng2024hybridflow} and score the student's top-16 token IDs with the three frozen teachers. The transfer objective combines the dense distillation reward with the student KL penalty. Answer graders are used only for evaluation. Implementation and training/evaluation configurations appear in Appendices~\ref{app:implementation} and~\ref{app:protocol}.

\subsection{Baselines and controls}
\label{sec:baselines}

We compare against the Base student, standard on-policy distillation (OPD) from the Final teacher \citep{agarwal2024gkd}, \opdtwo{} with its
sampled-token objective \citep{heo2026opd2}, and endpoint \directopd{} using $\log\teacherpost-\log\teacherpre$ \citep{feng2026directopd}. All methods share training and evaluation settings, except for method-specific hyperparameters.

Three controls on Qwen3-8B math retain centering and alter one component of the full method. \textbf{No gate} sets $g_t=1$ while retaining the gate-mean normalization rule. \textbf{Shuffled gate} permutes gates among valid response
prefixes within a batch, retaining each prefix's candidate weights and centered endpoint signal. \textbf{No normalization} fixes $\kappa_B=1$ while retaining the positive-cosine gate. Appendix~\ref{app:ablation-plan} gives definitions and invariants of the shuffle control. We separately vary the math Waypoint over teacher RL steps 500, 600, and 700 on Qwen3-8B in Appendix~\ref{app:waypoint-sensitivity}.

\section{Results and Analysis}
\label{sec:results}

We compare transfer performance across math and code
at three student scales (Section~\ref{sec:main-results}). We examine learning trajectories and performance relative to the Final teacher on Qwen3-8B math (Section~\ref{sec:training-dynamics}). We assess the roles of temporal gating, gate assignment, and mean normalization through ablations (Section~\ref{sec:ablation-results}).

\subsection{Main results across domains and scales}
\label{sec:main-results}

Tables~\ref{tab:full-math-results} and~\ref{tab:code-summary} report math scores and code averages. Full code results appear in
Appendix~\ref{app:full-code-results}. \method{} achieves the highest domain-average score among the evaluated student methods in all six settings.

% Generated by scripts/build_result_tables.py; edit data/main_results.json.
\begin{table}[!t]
  \caption{\textbf{Mathematical reasoning across three student sizes.} \texttt{mean@32} scores (\%) on seven benchmarks and their average. Bold marks the best score per student. Teacher checkpoints are references.}
  \label{tab:full-math-results}
  \centering
  \begingroup
  \small
  \setlength{\tabcolsep}{2.4pt}
  \renewcommand{\arraystretch}{1.12}
  % Zero-width group boxes keep the spanning bands from widening the Avg. column.
  \begin{tabular*}{\linewidth}{@{\extracolsep{\fill}}l*{8}{c}@{}}
    \toprule
    Method & \multicolumn{1}{c}{AIME24} & \multicolumn{1}{c}{AIME25} & \multicolumn{1}{c}{AIME26} & \multicolumn{1}{c}{\begin{tabular}[c]{@{}c@{}}HMMT25\\Feb\end{tabular}} & \multicolumn{1}{c}{\begin{tabular}[c]{@{}c@{}}HMMT25\\Nov\end{tabular}} & \multicolumn{1}{c}{\begin{tabular}[c]{@{}c@{}}HMMT26\\Feb\end{tabular}} & \multicolumn{1}{c}{AMC23} & \multicolumn{1}{c@{}}{Avg.} \\
    \midrule
    \multicolumn{9}{@{}c@{}}{\makebox[0pt][c]{\colorbox{black!6}{\makebox[\dimexpr\linewidth-2\fboxsep\relax][l]{\strut\textbf{Teacher: Qwen3-4B}}}}} \\
    Initial & 21.67 & 21.67 & 21.67 & 12.50 & 10.00 & 15.15 & 71.25 & 24.84 \\
    Waypoint & 58.96 & 54.79 & 56.77 & 31.35 & 35.73 & 31.72 & 93.44 & 51.82 \\
    Final & 65.83 & 60.00 & 60.83 & 35.83 & 40.83 & 35.13 & 94.45 & 56.13 \\
    \addlinespace[3pt]
    \multicolumn{9}{@{}c@{}}{\makebox[0pt][c]{\colorbox{black!6}{\makebox[\dimexpr\linewidth-2\fboxsep\relax][l]{\strut\textbf{Student: Qwen3-8B}}}}} \\
    Base & 29.17 & 20.83 & 18.33 & 11.67 & 13.33 & 18.94 & 70.00 & 26.04 \\
    OPD & 63.65 & 49.69 & 54.90 & 29.06 & 38.65 & 32.58 & 88.75 & 51.04 \\
    \opdtwo{} & 64.48 & 48.33 & 53.96 & 28.44 & 40.52 & 31.53 & \textbf{92.97} & 51.46 \\
    \directopd{} & 67.29 & 55.31 & 57.40 & 31.77 & 40.83 & 33.71 & 91.41 & 53.96 \\
    \addlinespace[1pt]
    \textbf{Waypoint-OPD} & \textbf{68.85} & \textbf{57.92} & \textbf{60.21} & \textbf{34.38} & \textbf{42.60} & \textbf{36.36} & 92.66 & \textbf{56.14} \\
    \addlinespace[3pt]
    \multicolumn{9}{@{}c@{}}{\makebox[0pt][c]{\colorbox{black!6}{\makebox[\dimexpr\linewidth-2\fboxsep\relax][l]{\strut\textbf{Student: Qwen3-14B}}}}} \\
    Base & 30.00 & 27.50 & 20.00 & 10.00 & 11.67 & 18.94 & 73.13 & 27.32 \\
    OPD & 69.17 & 55.52 & 58.44 & 36.77 & 45.21 & 31.82 & \textbf{96.25} & 56.17 \\
    \opdtwo{} & 69.48 & 46.98 & 54.38 & 33.13 & 41.98 & 31.91 & 91.95 & 52.83 \\
    \directopd{} & 70.10 & 56.46 & 61.98 & 37.92 & \textbf{47.60} & 35.13 & 94.06 & 57.61 \\
    \addlinespace[1pt]
    \textbf{Waypoint-OPD} & \textbf{70.21} & \textbf{60.52} & \textbf{63.75} & \textbf{40.21} & 47.08 & \textbf{36.55} & 96.09 & \textbf{59.20} \\
    \addlinespace[3pt]
    \multicolumn{9}{@{}c@{}}{\makebox[0pt][c]{\colorbox{black!6}{\makebox[\dimexpr\linewidth-2\fboxsep\relax][l]{\strut\textbf{Student: Qwen3-30B-A3B}}}}} \\
    Base & 31.67 & 22.50 & 22.50 & 10.00 & 13.33 & 21.21 & 75.00 & 28.03 \\
    OPD & 63.54 & 49.06 & 57.50 & 30.83 & 37.71 & 28.03 & 91.88 & 51.22 \\
    \opdtwo{} & 64.90 & 51.77 & 60.42 & 31.56 & 38.54 & 30.21 & 91.33 & 52.68 \\
    \directopd{} & 69.17 & 54.69 & 61.98 & 38.96 & 41.56 & \textbf{33.05} & 92.34 & 55.96 \\
    \addlinespace[1pt]
    \textbf{Waypoint-OPD} & \textbf{70.94} & \textbf{58.44} & \textbf{63.65} & \textbf{39.90} & \textbf{44.69} & 32.77 & \textbf{94.53} & \textbf{57.85} \\
    \bottomrule
  \end{tabular*}
  \endgroup
\end{table}

\paragraph{Math.}
\method{} improves over \directopd{}, the strongest average baseline for all three students, by 2.18, 1.59, and 1.89 percentage points for Qwen3-8B, Qwen3-14B, and Qwen3-30B-A3B, respectively. It improves 19 of 21 benchmark comparisons. The exceptions are HMMT25N for 14B (47.60 to 47.08) and HMMT26F for 30B-A3B (33.05 to 32.77). These results demonstrate improvements over endpoint transfer across student scales and both
dense and MoE architectures. The 14B and 30B-A3B students also exceed the Final teacher's 56.13 average by 3.07 and 1.72 points, respectively, while 8B essentially matches it at 56.14.
The 14B student surpasses the teacher on all seven math benchmarks.

\begin{table}[!t]
  \caption{\textbf{Average code reasoning performance.} Average \texttt{mean@8} scores (\%) over five evaluations. Qwen3-4B teacher checkpoints are shared references. Bold marks the best score per student. Full results: Appendix~\ref{app:full-code-results}.}
  \label{tab:code-summary}
  \centering
  \begingroup
  \small
  \setlength{\tabcolsep}{4pt}
  \renewcommand{\arraystretch}{1.12}
  \begin{tabularx}{0.88\linewidth}{@{}l*{3}{>{\centering\arraybackslash}X}@{}}
    \toprule
    \multicolumn{4}{@{}c@{}}{\textit{Teacher reference: Qwen3-4B}} \\
    \multicolumn{4}{@{}c@{}}{\textbf{Initial: 37.64 \qquad Waypoint: 52.22 \qquad Final: 55.37}} \\
    \midrule
    Method & Qwen3-8B & Qwen3-14B & Qwen3-30B-A3B \\
    \midrule
    Base & 41.03 & 47.72 & 50.33 \\
    OPD & 43.91 & 51.66 & 51.91 \\
    \opdtwo{} & 51.21 & 55.60 & 55.96 \\
    \directopd{} & 50.59 & 56.30 & 56.35 \\
    \addlinespace[1pt]
    \textbf{Waypoint-OPD} & \textbf{53.25} & \textbf{57.82} & \textbf{57.54} \\
    \bottomrule
  \end{tabularx}
  \endgroup
\end{table}

\paragraph{Code.}
\method{} improves over \directopd{} by 2.66, 1.52, and 1.19 percentage points on average for Qwen3-8B, Qwen3-14B, and Qwen3-30B-A3B, respectively, with gains in all 15 code comparisons. \opdtwo{} achieves higher scores on 14B LCB v5 and on 30B-A3B MBPP and MBPP+, while \method{} retains the highest average for each student (Appendix~\ref{app:full-code-results}). The 14B and 30B-A3B students also surpass the Final teacher's 55.37 average
by 2.45 and 2.17 points, respectively, each scoring higher on four of the five code evaluations.

\paragraph{Interpreting the endpoint comparison.}
The gains over \directopd{} show that \method{} improves transfer performance across both domains and all three student scales. OPD and \opdtwo{} provide complementary comparisons with Final-teacher imitation and selective distillation based on advantage sign agreement, respectively.

\subsection{Learning dynamics}
\label{sec:training-dynamics}

\begin{figure}[!htbp]
  \centering
  \includegraphics[width=\linewidth]{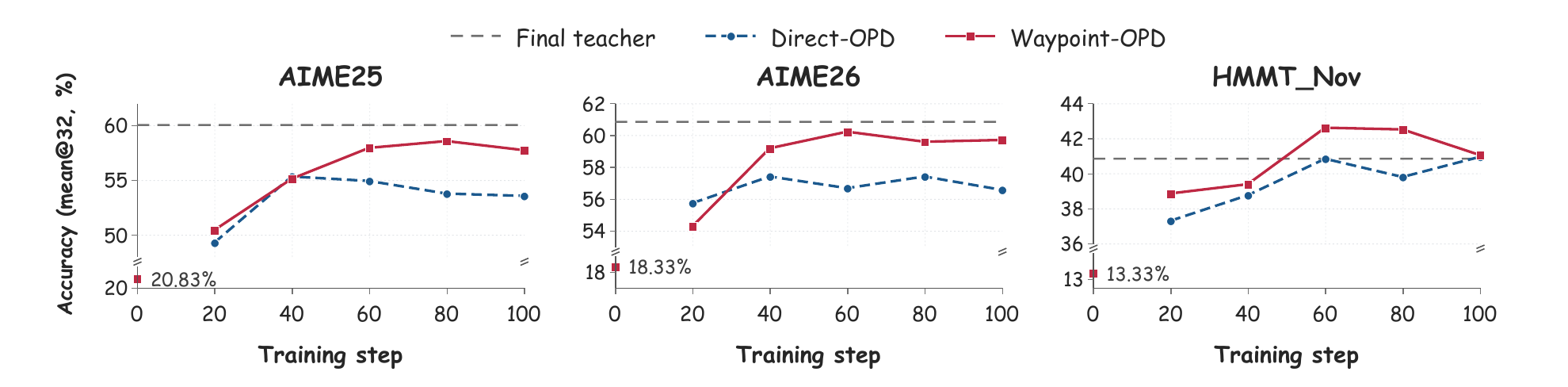}
  \caption{\textbf{Evaluation accuracy during training.} Qwen3-8B results on AIME 2025, AIME 2026, and HMMT November 2025 at steps 0, 20, 40, 60, 80, and 100. Scores are \texttt{mean@32} percentages. Solid red curves denote \method{} and dashed blue curves denote \directopd{}. Gray dashed lines show the Final Qwen3-4B teacher scores.}
  \label{fig:evaluation-accuracy}
\end{figure}

Figure~\ref{fig:evaluation-accuracy} compares \method{} and \directopd{} at matched evaluation steps during Qwen3-8B math training. Unlike the selected-checkpoint results in Table~\ref{tab:full-math-results}, these trajectories show the performance gap over the same 100 updates.

\paragraph{Consistent gains across training checkpoints.}
\method{} delivers consistent gains over \directopd{} across both benchmarks and training checkpoints. Excluding the shared initialization, it scores higher in 13 of the 15 benchmark--checkpoint comparisons, with two exceptions early in training. From step 60 onward, \method{} leads on all three benchmarks at every recorded evaluation, showing that its advantage is sustained through the end of training. The step-100 gains are 4.17, 3.13, and 0.10 percentage points on AIME 2025, AIME 2026, and HMMT November 2025, respectively. These results show that gains extend beyond the selected checkpoint to sustained advantages during training.

\paragraph{Comparison with the Final teacher.}
The gray reference lines distinguish gains over endpoint transfer from performance relative to the teacher. On HMMT November 2025, \method{} exceeds the Final teacher's 40.83 at steps 60, 80, and 100, reaching 42.60 at step 60. On both AIME benchmarks, its recorded scores remain below the corresponding teacher reference. Improvements over \directopd{} therefore
coexist with different teacher--student orderings across benchmarks. The sustained HMMT advantage demonstrates that \method{} can use the teacher's policy shifts to achieve performance beyond the Final teacher's score.

\subsection{Gate and normalization ablations}
\label{sec:ablation-results}

% Generated by scripts/build_result_tables.py; edit data/ablation_results.json.
\begin{table}[!htbp]
  \caption{\textbf{Gate and normalization ablations on Qwen3-8B mathematical reasoning.} Bold marks the best score in each column.}
  \label{tab:gate-ablation}
  \centering
  \begingroup
  \small
  \setlength{\tabcolsep}{2.4pt}
  \renewcommand{\arraystretch}{1.12}
  \begin{tabular*}{\linewidth}{@{\extracolsep{\fill}}l*{8}{c}@{}}
    \toprule
    Method & \multicolumn{1}{c}{AIME24} & \multicolumn{1}{c}{AIME25} & \multicolumn{1}{c}{AIME26} & \multicolumn{1}{c}{\begin{tabular}[c]{@{}c@{}}HMMT25\\Feb\end{tabular}} & \multicolumn{1}{c}{\begin{tabular}[c]{@{}c@{}}HMMT25\\Nov\end{tabular}} & \multicolumn{1}{c}{\begin{tabular}[c]{@{}c@{}}HMMT26\\Feb\end{tabular}} & \multicolumn{1}{c}{AMC23} & \multicolumn{1}{c@{}}{Avg.} \\
    \midrule
    No gate & 66.25 & 54.27 & 56.67 & 32.60 & 39.79 & 34.66 & 90.23 & 53.50 \\
    Shuffled gate & 63.33 & 52.92 & 54.58 & 30.00 & 37.08 & 31.06 & 89.45 & 51.20 \\
    No normalization & 68.13 & 56.46 & 58.33 & \textbf{34.79} & 41.56 & 35.23 & 92.03 & 55.22 \\
    \directopd{} & 67.29 & 55.31 & 57.40 & 31.77 & 40.83 & 33.71 & 91.41 & 53.96 \\
    \addlinespace[2pt]
    \textbf{Waypoint-OPD} & \textbf{68.85} & \textbf{57.92} & \textbf{60.21} & 34.38 & \textbf{42.60} & \textbf{36.36} & \textbf{92.66} & \textbf{56.14} \\
    \bottomrule
  \end{tabular*}
  \endgroup
\end{table}

\paragraph{Removing the gate.}
On Qwen3-8B math, removing the gate lowers Avg. from 56.14 to 53.50, a decrease of 2.64 points, with lower scores on all seven benchmarks (Table~\ref{tab:gate-ablation}). Retaining centering and normalization without consistency-dependent weighting therefore does not recover the full method's gain. This comparison does not isolate the effect of centering.

\paragraph{Disrupting the gate--prefix association.}
Shuffling gates lowers Avg. to 51.20, which is 4.94 points below the full method and 2.30 below the ungated variant, with the same ordering on every benchmark. Random reassignment is thus more detrimental than removing the selective weighting in this setting, supporting the value of matching each gate to the prefix whose temporal agreement produced it.

\paragraph{Removing mean normalization.}
Fixing $\kappa_B=1$ while retaining the positive-cosine gate lowers Avg. by 0.92 points to 55.22. Normalization improves six of seven benchmarks. The exception is HMMT February 2025 (34.79 without normalization versus 34.38 with it). The unnormalized variant still exceeds \directopd{} by 1.26 points on average and on all seven benchmarks. Thus, the advantage persists without
batch normalization, which provides an additional average benefit here. The remaining gain cannot be assigned entirely to gating because centering and other differences from \directopd{} are also retained.

\paragraph{Ablation insights.}
These ablations support temporal weighting, its prefix assignment, and an incremental benefit from normalization in Qwen3-8B math. Their gaps are conditional on the retained components, not independent additive contributions. For a fixed rollout batch, shuffling preserves the gate distribution and
$\kappa_B$, while advantage and gradient norms depend additionally on the gate--signal pairing (Appendix~\ref{app:ablation-plan}). Supplementary diagnostics show stronger average gate attenuation in later batches, while mean-gate normalization does not fix the reward norm
(Appendix~\ref{app:gate_diagnostics}).

\section{Related Work}
\label{sec:related-work}

\paragraph{On-policy distillation and policy-shift transfer.}
Knowledge distillation transfers teacher distributions or generated sequences \citep{hinton2015distilling,kim2016sequence}, while on-policy variants supervise student-generated trajectories \citep{agarwal2024gkd}. Complementary work refines divergence objectives and the use of teacher- and student-generated
data \citep{wen2023fdistill,gu2024minillm,ko2024distillm,ko2025distillm2}. Weak-to-strong generalization asks whether weaker supervision can elicit stronger capabilities \citep{burns2024w2s}. WSPO and proxy-tuning reuse a weak
model's adaptation contrast for training and decoding, respectively \citep{zhu2025wspo,liu2024proxytuning}.
Within on-policy training, \directopd{} transfers an RL-induced policy change through an endpoint log-ratio on student rollouts \citep{feng2026directopd}. W2S-OPD constructs a proxy teacher by adding a weak-model logit contrast to student base logits, then minimizes reverse KL on student rollouts. Its contrasts span RL checkpoints, model scales, and correct versus incorrect
hints \citep{yu2026w2sopd}. OPRD uses a centered teacher--reference shift to amplify the component of a verifier-driven gradient along that direction \citep{park2026oprd}. \method{} builds on \directopd{}, weighting local endpoint signals by
agreement between shifts before and after a Waypoint checkpoint.

\paragraph{Selective supervision in distillation.}
Teacher--student compatibility affects the utility of on-policy supervision \citep{li2026rethinkingopd}. Prune-OPD uses support overlap for suffix attenuation \citep{yang2026pruneopd}. EOPD uses teacher uncertainty in its entropy-dependent forward KL \citep{jin2026entropyopd}, while PW-OPSD uses token position in its reference-conditioned self-distillation \citep{liu2026pwopsd}. \opdtwo{} gates sampled-token updates by sign agreement between centered teacher--base and teacher--student advantages \citep{heo2026opd2}, whereas RG-OPD gates trajectories using agreement between verifier-derived advantages and teacher--student sequence log-likelihood gaps \citep{akhondzadeh2026rgopd}. \method{} uses different evidence: consecutive shifts in the teacher's own RL trajectory over student-supported candidate actions. Their weighted directional agreement produces a prefix-level gate for the dense endpoint
update, without verifier feedback during transfer. This agreement is a weighting criterion, not a guarantee of correctness.

\paragraph{Checkpoint selection and progressive distillation.}
A teacher's final performance need not determine its teaching utility: early stopping and intermediate checkpoint selection can improve distillation \citep{cho2019efficacy,wang2022checkpointkd}. Pro-KD and progressive distillation use successive teacher checkpoints
as targets or a curriculum \citep{rezagholizadeh2021prokd,panigrahi2025progressive}.
\directopd{} also studies RL checkpoint choice for transfer
\citep{feng2026directopd}. For fixed Initial and Final endpoints, Waypoint measures directional consistency between the two shifts. It supplies neither an imitation target nor a curriculum. The centered Initial-to-Final shift remains the transferred object.

\section{Conclusion}
\label{sec:conclusion}

We presented \method{}, which uses agreement between consecutive teacher shifts to transfer RL-induced policy changes. An intermediate Waypoint checkpoint reveals whether early and late changes in relative action preferences reinforce or oppose each other at student-generated prefixes. This information, unavailable from endpoint differences alone, determines the weight of the centered Initial-to-Final shift in the student's on-policy objective. The method requires one additional frozen-teacher scoring pass and no verifier calls during transfer.
Experiments on mathematical and code reasoning show gains over endpoint transfer across three Qwen3 student scales. Ablations on Qwen3-8B math support prefix-specific gate assignment and batch-scale normalization. These findings suggest that intermediate teacher checkpoints provide evidence for weighting local policy changes during student learning.

\subsection*{AI use statement}

We use generative AI tools for language polishing, assistance in clarifying mathematical derivations, and consistency checks of notation and exposition. These tools were not used for experimental design, implementation, or execution. We take full responsibility for the final content of this work.

\subsection*{Ethics statement}

\paragraph*{Use of Human Annotations}
Human annotations are only used in methodological research at the beginning of the work, to assist in analyzing the feasibility of the proposed solution. Annotators consented to the use of data for research purposes. We ensure that the privacy of all annotators is protected throughout the annotation process, and all of them are adequately paid according to local standards. Human annotations are not applied during the evaluation of our method.
\paragraph*{Risks}
In this paper, all datasets are obtained from official sources. The datasets adopted have been anonymized and do not contain offensive information. However, we cannot guarantee that the datasets do not contain socially harmful or toxic language.

\subsection*{Reproducibility statement}

The transfer objective is specified in Section~\ref{sec:method}, with supporting derivations and signal properties in Appendix~\ref{app:proofs} and pseudocode and
optimization details in Appendix~\ref{app:implementation}.
Section~\ref{sec:experimental-setup} describes the teacher checkpoints, student models, training and evaluation data, evaluation metrics, and student checkpoint selection. Appendix~\ref{app:protocol} reports hyperparameters, hardware, and evaluation settings.

Baselines and controls are described in Section~\ref{sec:baselines}, with precise definitions of the gate and normalization ablations in Appendix~\ref{app:ablation-plan} and the Waypoint-position ablation setup in Appendix~\ref{app:waypoint-sensitivity}. Per-benchmark results and domain averages appear in Tables~\ref{tab:full-math-results}
and~\ref{tab:full-code-results}, and gate and normalization ablation results appear in Table~\ref{tab:gate-ablation}.

\bibliography{iclr2027_conference}
\bibliographystyle{iclr2027_conference}

\newpage
\appendix

\renewcommand{\floatpagefraction}{0.8}

\section{Derivations and Signal Properties}
\label{app:proofs}

\subsection{Endpoint preservation and gate properties}
\label{app:endpoint-preservation}

At fixed $s_t$, $\cK_t$, and $q_t$, with finite teacher log scores, linearity of centering gives
\begin{equation}
  u_t+v_t=C_{q_t}[\delta^{(1)}_t+\delta^{(2)}_t]
  =C_{q_t}[\log\teacherpost-\log\teacherpre]=d_t.
  \label{eq:waypoint-endpoint-preservation}
\end{equation}
Hence Waypoint affects $g_t$ and $\kappa_B$, while the transferred endpoint $d_t$ remains fixed. This comparison holds at fixed student prefixes and weights; subsequent training trajectories can differ.

The remaining properties follow directly from the definitions.
First, $C_{q_t}[z+b]=C_{q_t}[z]$ for any action-independent constant $b$, so such offsets in teacher scores leave the transfer signal unchanged. Second, outside the low-norm fallback, the gate is one for positively collinear shifts and zero for nonpositive weighted cosine. Independent positive rescaling of $u_t$ and $v_t$ leaves this gate unchanged provided neither segment enters the fallback; their sum $d_t$ can change.
Finally, $g_t\in[0,1]$ and $\kappa_B>0$ ensure that $A_t(a)$ cannot reverse the sign of $d_t(a)$. The effective multiplier $\kappa_Bg_t$ can exceed one: batch normalization targets the mean gate, not a fixed reward norm. These are properties of the construction; the usefulness of temporal weighting is evaluated empirically.

\subsection{Zero-sum coefficients and the local gradient}
\label{app:local-gradient}

Centering under $q_t$ immediately gives
\begin{equation}
  \sum_{a\in\cK_t}A_t(a)
  =\kappa_Bg_t\sum_{a\in\cK_t}q_t(a)d_t(a)=0.
  \label{eq:zero-sum-coefficients}
\end{equation}
With $\cK_t$ fixed and $A_t$ detached, this cancels the common softmax term. Writing $z_b$ for the student logit of vocabulary token $b$ at $s_t$,
\begin{equation}
  \frac{\partial}{\partial z_b}
    \left[-\sum_{a\in\cK_t}A_t(a)\log\student(a\mid s_t)\right]
  =\begin{cases}
    -A_t(b), & b\in\cK_t,\\
    0, & b\notin\cK_t.
  \end{cases}
  \label{eq:zero-sum-logit-gradient}
\end{equation}
The PPO surrogate has this gradient at the rollout policy
(Appendix~\ref{app:ppo-surrogate}). This local result excludes the KL term and does not imply that a shared-parameter update preserves candidate probability mass or leaves noncandidate probabilities unchanged.

\section{Algorithm and Implementation Details}
\label{app:implementation}

\begin{algorithm}[H]
\caption{Agreement-gated on-policy distillation of RL-induced policy shifts}
\label{alg:waypoint-opd}
\begin{algorithmic}[1]
\REQUIRE Prompts $\mathcal{D}$, student and reference policies $\student,\studentref$, frozen Initial, Waypoint, and Final teachers $\teacherpre,\teachermid,\teacherpost$, hyperparameters $k,\tau,\varepsilon$, scale bounds, and KL coefficient $\beta$.
\FOR{each update}
  \STATE Sample prompts $x$ and responses
  $y\sim\student(\cdot\mid x)$.
  \FOR{each valid prefix $s_t=(x,y_{<t})$}
    \STATE Obtain $\cK_t=\operatorname{TopK}(\student(\cdot\mid s_t),k)$ and detached normalized weights $q_t$.
    \STATE Score $\cK_t$ with $\teacherpre$, $\teachermid$, and $\teacherpost$.
    \STATE Form $u_t=C_{q_t}[\log\teachermid-\log\teacherpre]$ and $v_t=C_{q_t}[\log\teacherpost-\log\teachermid]$.
    \STATE Compute $c_t$ with Eq.~\ref{eq:weighted-cosine} and set $g_t=\max(0,c_t)$. Use $g_t=1$ if either centered norm is at most $\varepsilon$.
    \STATE Set $d_t=u_t+v_t$.
  \ENDFOR
  \STATE Compute $\kappa_B$ with Eq.~\ref{eq:gate-normalization}.
  \STATE Set $A_t(a)=\sg[\kappa_B\,q_t(a)\,g_t\,d_t(a)]$.
  \STATE Update $\student$ using the PPO surrogate in Eq.~\ref{eq:dual-clip} and the KL penalty in Eq.~\ref{eq:waypoint-objective}.
\ENDFOR
\end{algorithmic}
\end{algorithm}

\subsection{Candidate-wise dual-clipped PPO}
\label{app:ppo-surrogate}

We use the candidate-wise dual-clipped PPO surrogate
\citep{schulman2017ppo,ye2020dualclip}. Let $\pi_{\text{old}}$ denote the student policy at rollout time, with $r_t(a)=\student(a\mid s_t)/\pi_{\text{old}}(a\mid s_t)$,
$\bar r_t(a)=\operatorname{clip}(r_t(a),1-\epsilon,1+\epsilon)$, and let $c>1$ be the dual-clip constant, distinct from $c_t$.
For each candidate, the maximized surrogate is
\begin{equation}
  J_t(a)=
  \begin{cases}
    \min\{r_t(a)A_t(a),\bar r_t(a)A_t(a)\}, & A_t(a)\geq0,\\
    \max\{\min[r_t(a)A_t(a),\bar r_t(a)A_t(a)],cA_t(a)\},
      & A_t(a)<0.
  \end{cases}
  \label{eq:dual-clip}
\end{equation}
The loss sums $-J_t(a)$ over candidates and uses token-mean aggregation over valid response tokens, together with the sampled-token KL penalty to $\studentref$ in Eq.~\ref{eq:waypoint-objective}. At the rollout policy, $r_t(a)=1$ and clipping is inactive, so maximizing $\sum_{a\in\cK_t}J_t(a)$ gives the same local distillation update as Eq.~\ref{eq:waypoint-objective} (Appendix~\ref{app:local-gradient}).

\subsection{Implementation conventions}

Candidate supports $\cK_t$ and detached coefficients $A_t(a)$ remain fixed during each student update. Batch-scale normalization uses valid response prefixes across the global rollout batch $B$. The low-norm fallback retains the endpoint signal when a segment has too little weighted variation to define a reliable direction, using $\varepsilon=10^{-8}$. The normalization floor $\varepsilon_{\text{num}}$ is the smallest positive normal float32 value ($\approx 1.18\times10^{-38}$), distinct from the low-norm threshold. Training and evaluation configurations appear in Appendix~\ref{app:protocol}.

\section{Full Experimental Protocol}
\label{app:protocol}

We use \texttt{verl} \citep{sheng2024hybridflow} with FSDP
\citep{zhao2023fsdp} for training and vLLM \citep{kwon2023vllm} for rollouts. Training and inference are performed on a single node with 8 NVIDIA H20 GPUs. Tables~\ref{tab:teacher-training-setup}, \ref{tab:training-setup}, and~\ref{tab:domain-config} summarize the teacher and transfer training hyperparameters, model configurations, and evaluation benchmarks.

\paragraph{Teacher training.}
For math RL training, we select approximately 57K examples with difficulty levels $\geq 6$ from DeepMath (DeepMath-103K) \citep{he2025deepmath}. For code RL training, we use Eurus-RL-Code \citep{cui2025process}, which contains approximately 25K training examples. We train both Qwen3-4B teachers with group relative policy optimization (GRPO) \citep{shao2024deepseekmath} for 1,200 optimization steps using the same hyperparameters in Table~\ref{tab:teacher-training-setup}; only the training datasets differ.

\begin{table}[!htbp]
  \caption{GRPO training configuration shared by the math and code teachers.}
  \label{tab:teacher-training-setup}
  \centering
  \small
  \setlength{\tabcolsep}{4.2pt}
  \begin{tabular}{lc}
    \toprule
    Setting & Value \\
    \midrule
    Train batch size & 128 \\
    Micro batch size & 128 \\
    Rollout $n$ & 8 \\
    Maximum prompt length & 2,048 \\
    Maximum response length & 16,384 \\
    Temperature & 1.0 \\
    Top-$p$ & 1.0 \\
    Learning rate & $10^{-6}$ \\
    Optimization steps & 1,200 \\
    KL coefficient & 0.0 \\
    \bottomrule
  \end{tabular}
\end{table}

\begin{table}[!htbp]
  \caption{Transfer configuration shared by math and code, except for the transfer datasets.}
  \label{tab:training-setup}
  \centering
  \small
  \setlength{\tabcolsep}{4.2pt}
  \begin{tabular}{lc}
    \toprule
    Setting & Value \\
    \midrule
    Math training dataset & DAPO-Math-17K \\
    Code training dataset & Codeforces-derived \\
    \midrule
    Train batch size & 256 \\
    Mini batch size & 64 \\
    Rollout $n$ & 1 \\
    Maximum prompt length & 2,048 \\
    Maximum response length & 16,384 \\
    Temperature & 1.0 \\
    Learning rate & $10^{-6}$ \\
    Optimization steps & 100 \\
    KL coefficient $\beta$ & 2.5 \\
    Student support $k$ & 16 \\
    Low-norm threshold $\varepsilon$ & $10^{-8}$ \\
    Target gate mean $\tau$ & 0.8 \\
    Scale clip & $[0.75,1.5]$ \\
    \bottomrule
  \end{tabular}
\end{table}

\paragraph{Evaluation settings.}
Math and code evaluation use the same rollout settings: temperature 0.7, top-$p$ 0.95, and a maximum generation length of 16,384 tokens. We report \texttt{mean@32} for math and \texttt{mean@8} for code, averaging correctness over 32 and eight responses per problem, respectively, and then over the problems in each benchmark. MBPP \citep{austin2021mbpp} and MBPP+ from EvalPlus \citep{liu2023evalplus} use the same problems with the original tests and the original plus additional tests, respectively. We also evaluate on LiveCodeBench v4, LiveCodeBench v5, and LiveCodeBench v6
\citep{jain2025livecodebench}. For both domains, we select each trained student's checkpoint by its validation accuracy on AIME 2024 and report all benchmark results from that checkpoint.

\begin{table}[H]
  \caption{Model and evaluation configuration.}
  \label{tab:domain-config}
  \centering
  \small
  \setlength{\tabcolsep}{4.2pt}
  \begin{tabularx}{\linewidth}{@{}>{\raggedright\arraybackslash}p{0.27\linewidth}
      >{\raggedright\arraybackslash}X@{}}
    \toprule
    Item & Value \\
    \midrule
    Initial teacher & Qwen3-4B \\
    Waypoint teacher & RL step 600 \\
    Final teacher & RL step 1200 \\
    Students & Qwen3-8B, Qwen3-14B, Qwen3-30B-A3B \\
    \midrule
    Math evaluation & AIME~2024, AIME~2025, AIME~2026, HMMT~February~2025,HMMT~November~2025, HMMT~February~2026, AMC~2023 \\
    \midrule
    Code evaluation & MBPP, MBPP+, LiveCodeBench~v4, LiveCodeBench~v5,LiveCodeBench~v6 \\
    \bottomrule
  \end{tabularx}
\end{table}

\section{Full Code Results}
\label{app:result-tables}
\label{app:full-code-results}

Table~\ref{tab:full-code-results} presents the complete code evaluation results corresponding to the averages summarized in
Table~\ref{tab:code-summary} in the main text. It includes the three teacher checkpoints and all evaluated student methods.

% Generated by scripts/build_result_tables.py; edit data/main_results.json.
\begin{table}[!htbp]
  \caption{\textbf{Full code reasoning results across three student sizes.} Scores are \texttt{mean@8} percentages. Avg. is the unweighted mean across the five evaluations. Bold marks the best score within each student block. Teacher checkpoints are reference rows.}
  \label{tab:full-code-results}
  \centering
  \begingroup
  \small
  \setlength{\tabcolsep}{2.4pt}
  \renewcommand{\arraystretch}{1.12}
  \begin{tabularx}{\linewidth}{@{}l*{6}{>{\centering\arraybackslash}X}@{}}
    \toprule
    Method & MBPP & MBPP+ & LCB v4 & LCB v5 & LCB v6 & Avg. \\
    \midrule
    \multicolumn{7}{@{}l@{}}{\colorbox{black!6}{\makebox[\dimexpr\linewidth-2\fboxsep\relax][l]{\strut\textbf{Teacher: Qwen3-4B}}}} \\
    Initial & 62.53 & 54.53 & 25.37 & 20.36 & 25.43 & 37.64 \\
    Waypoint & 83.33 & 70.37 & 36.01 & 37.20 & 34.21 & 52.22 \\
    Final & 88.00 & 72.65 & 38.61 & 40.57 & 37.00 & 55.37 \\
    \addlinespace[3pt]
    \multicolumn{7}{@{}l@{}}{\colorbox{black!6}{\makebox[\dimexpr\linewidth-2\fboxsep\relax][l]{\strut\textbf{Student: Qwen3-8B}}}} \\
    Base & 67.16 & 58.63 & 30.82 & 23.35 & 25.21 & 41.03 \\
    OPD & 69.58 & 57.67 & 35.64 & 26.95 & 29.71 & 43.91 \\
    \opdtwo{} & 79.10 & 68.68 & 36.54 & 37.72 & 34.00 & 51.21 \\
    \directopd{} & 78.14 & 65.81 & 36.63 & 37.95 & 34.43 & 50.59 \\
    \addlinespace[1pt]
    \textbf{Waypoint-OPD} & \textbf{82.97} & \textbf{69.38} & \textbf{37.87} & \textbf{40.19} & \textbf{35.86} & \textbf{53.25} \\
    \addlinespace[3pt]
    \multicolumn{7}{@{}l@{}}{\colorbox{black!6}{\makebox[\dimexpr\linewidth-2\fboxsep\relax][l]{\strut\textbf{Student: Qwen3-14B}}}} \\
    Base & 83.86 & 71.69 & 30.94 & 26.95 & 25.14 & 47.72 \\
    OPD & 84.49 & 72.62 & 35.64 & 33.53 & 32.00 & 51.66 \\
    \opdtwo{} & 85.98 & 72.92 & 38.61 & \textbf{43.56} & 36.93 & 55.60 \\
    \directopd{} & 85.32 & 73.58 & 43.44 & 42.96 & 36.21 & 56.30 \\
    \addlinespace[1pt]
    \textbf{Waypoint-OPD} & \textbf{87.57} & \textbf{74.60} & \textbf{45.54} & 43.11 & \textbf{38.29} & \textbf{57.82} \\
    \addlinespace[3pt]
    \multicolumn{7}{@{}l@{}}{\colorbox{black!6}{\makebox[\dimexpr\linewidth-2\fboxsep\relax][l]{\strut\textbf{Student: Qwen3-30B-A3B}}}} \\
    Base & 86.34 & 73.18 & 32.55 & 28.67 & 30.93 & 50.33 \\
    OPD & 86.61 & 72.25 & 35.15 & 32.49 & 33.07 & 51.91 \\
    \opdtwo{} & \textbf{88.23} & \textbf{74.44} & 40.10 & 41.62 & 35.43 & 55.96 \\
    \directopd{} & 87.24 & 73.77 & 42.33 & 41.54 & 36.86 & 56.35 \\
    \addlinespace[1pt]
    \textbf{Waypoint-OPD} & 88.00 & 74.21 & \textbf{43.94} & \textbf{42.89} & \textbf{38.64} & \textbf{57.54} \\
    \bottomrule
  \end{tabularx}
  \endgroup
\end{table}

\section{Component Ablation Details}
\label{app:ablation-plan}

We detail the three controls evaluated on Qwen3-8B mathematical reasoning in Table~\ref{tab:gate-ablation}. All controls retain $q_t$-centering and the student KL penalty and use the training and evaluation settings in Appendix~\ref{app:protocol}. Each control collects rollouts from its current student policy.

\paragraph{No gate.}
This control sets $g_t=1$ at every valid response prefix while retaining the centered endpoint signal and the gate-mean normalization rule in Eq.~\ref{eq:gate-normalization}. With the default settings, the rule reduces to the constant multiplier
\[
  \kappa_B=\operatorname{clip}(0.8,0.75,1.5)=0.8.
\]
This control removes consistency-dependent weighting while retaining centering and the normalization rule.

\paragraph{Shuffled gate.}
We compute gates using the standard rule, including the low-norm fallback, and randomly permute them among valid response prefixes in the global rollout batch $B$. For a permutation $\sigma$, the advantage is
\[
  A_t^{\text{shuffle}}(a)
  =\sg[\kappa_B q_t(a)g_{\sigma(t)}d_t(a)].
\]
Candidate weights $q_t$ and endpoint signals $d_t$ remain at their original prefixes. For a fixed rollout batch, the permutation preserves the gate distribution and its mean, leaving $\kappa_B$ unchanged. It breaks the association between each prefix and its own temporal-agreement weight. Advantage and gradient norms depend additionally on this gate--signal pairing.

\paragraph{No normalization.}
This control fixes $\kappa_B=1$ while retaining centering, the positive-cosine gate, and the low-norm fallback:
\[
  A_t^{\text{unnorm}}(a)
  =\sg[q_t(a)g_t d_t(a)].
\]
For a fixed batch, removing the common multiplier preserves relative gate weights across prefixes while removing the batch-dependent rescaling of the distillation signal relative to the retained KL penalty.

\section{Sensitivity to Waypoint Checkpoint Selection}
\label{app:waypoint-sensitivity}

\paragraph{Experimental setup.}
We examine whether transfer performance depends on the precise position of the intermediate teacher checkpoint. On Qwen3-8B mathematical reasoning, we replace the default Waypoint at teacher RL step 600 with checkpoints at steps 500 and 700 from the same Qwen3-4B RL run. The Initial teacher before RL and the Final teacher at step 1200 remain fixed. The other training and evaluation settings follow the default Qwen3-8B math setup in
Appendix~\ref{app:protocol}. Table~\ref{tab:waypoint-ablation}
reports all seven benchmark scores and their unweighted average.

\begin{table}[!htbp]
  \caption{\textbf{Sensitivity to the Waypoint checkpoint on Qwen3-8B mathematical reasoning.} Only the intermediate teacher checkpoint is varied, while the Initial and Final teachers are fixed. Step numbers refer to the teacher's RL training. Bold marks the best score in each column.}
  \label{tab:waypoint-ablation}
  \centering
  \begingroup
  \small
  \setlength{\tabcolsep}{2.4pt}
  \renewcommand{\arraystretch}{1.12}
  \begin{tabular*}{\linewidth}{@{\extracolsep{\fill}}l*{8}{c}@{}}
    \toprule
    Waypoint & AIME24 & AIME25 & AIME26 & \begin{tabular}[c]{@{}c@{}}HMMT25\\Feb\end{tabular} & \begin{tabular}[c]{@{}c@{}}HMMT25\\Nov\end{tabular} & \begin{tabular}[c]{@{}c@{}}HMMT26\\Feb\end{tabular} & AMC23 & Avg. \\
    \midrule
    Step 500 & 69.06 & \textbf{58.13} & 59.79 & 33.96 & 42.60 & \textbf{36.74} & \textbf{92.97} & \textbf{56.18} \\
    Step 600 & 68.85 & 57.92 & \textbf{60.21} & 34.38 & 42.60 & 36.36 & 92.66 & 56.14 \\
    Step 700 & \textbf{69.27} & 57.60 & 59.79 & \textbf{34.58} & \textbf{42.81} & 36.36 & 92.50 & 56.13 \\
    \bottomrule
  \end{tabular*}
  \endgroup
\end{table}

\paragraph{Small changes in aggregate performance.}
The average scores are 56.18, 56.14, and 56.13 for steps 500, 600, and 700, respectively, giving a total range of only 0.05 percentage points. Relative to the default step 600, moving the Waypoint earlier changes Avg. by $+0.04$ points, and moving it later changes Avg. by $-0.01$ points. All three choices exceed the \directopd{} average of 53.96 by 2.17--2.22 points and outperform it on every individual benchmark (Table~\ref{tab:full-math-results}). Thus, the observed transfer advantage is preserved across these neighboring Waypoints.

\paragraph{Stability across Waypoint choices.}
Across the three tested Waypoints, the score range on each benchmark is at most 0.62 percentage points, with the largest range on HMMT25 February (33.96--34.58). Together with the 0.05-point range in Avg. and the consistent gains over \directopd{}, these results show stable transfer performance across the tested neighboring Waypoints on Qwen3-8B math.

\section{Gate Diagnostics}
\label{app:gate_diagnostics}

\begin{figure}[!htbp]
  \centering
  \includegraphics[width=\linewidth]{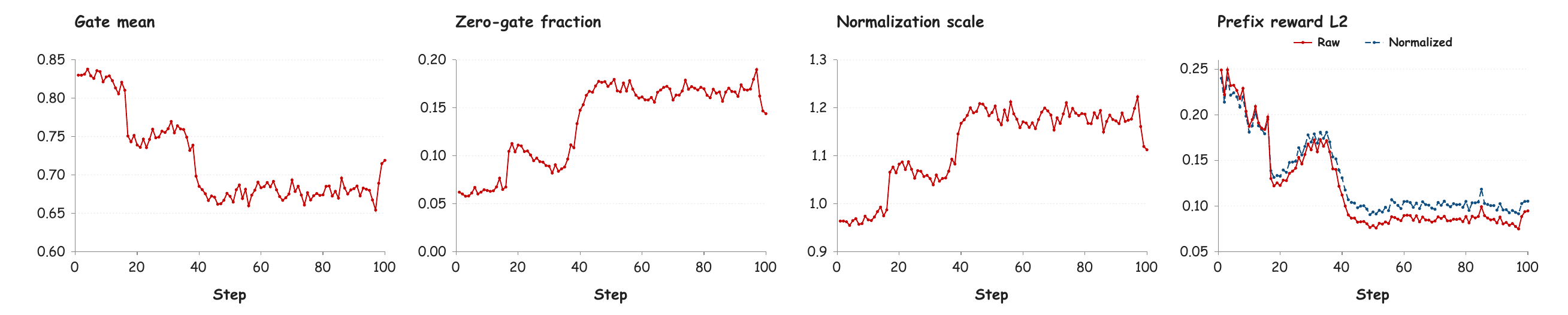}
  \caption{\textbf{Gate diagnostics on Qwen3-8B mathematical reasoning.} Statistics for \method{} over 100 training steps. From left to right: mean gate $g_t$, zero-gate fraction, batch-scale normalization multiplier $\kappa_B$, and mean per-prefix reward-vector L2 norm. Raw uses the gated, student-weighted endpoint signal $q_t(a)g_td_t(a)$; Normalized uses the
  final candidate coefficients $A_t(a)$ in Eq.~\ref{eq:advantage} from the same run. All statistics aggregate valid response prefixes.}
  \label{fig:gate-diagnostics}
\end{figure}

Figure~\ref{fig:gate-diagnostics} examines agreement gating and batch-scale normalization over 100 training steps of \method{} on Qwen3-8B mathematical reasoning, using the configuration in Appendix~\ref{app:protocol}.

\paragraph{Temporal weighting over training.}
The mean gate $g_t$ decreases from 0.810 in the first 20 steps to 0.683 in the last 20, while the zero-gate fraction increases from 7.26\% to 16.59\%. These summaries are unweighted averages of step-level statistics. The gate therefore applies stronger average attenuation before batch-scale normalization in later batches and removes the centered endpoint signal at a larger fraction of valid response prefixes. With the Initial, Waypoint, and Final teachers frozen, these statistics track temporal agreement on the evolving student-generated prefixes and candidate distributions.

\paragraph{Batch-scale normalization.}
The multiplier $\kappa_B$ ranges from 0.955 to 1.223, remaining within the configured bounds $[\kappa_{\text{min}},\kappa_{\text{max}}]=[0.75,1.5]$.
It is below one for the first 16 steps and above one thereafter,
maintaining the mean rescaled gate at the target $\tau=0.8$ to numerical precision. As specified in Eq.~\ref{eq:gate-normalization}, this shared positive multiplier preserves relative gate weights and zero gates. It also scales each prefix's reward-vector L2 norm by $\kappa_B$. Over the last 20 steps, the mean per-prefix norms average 0.0855 before scaling and 0.1002 after scaling. Both norms decrease overall during training, while the mean rescaled gate remains stable. These diagnostics show how batch-scale normalization maintains the average gate scale as temporal weighting adapts to student-generated prefixes. The performance benefit is reflected in Table~\ref{tab:gate-ablation}: mean normalization improves the Qwen3-8B math average from 55.22 to 56.14.

\section{Training Time and Response Length}
\label{app:efficiency}

Table~\ref{tab:efficiency} compares the total training time over 100 training steps of Qwen3-8B on math using eight H20 GPUs, excluding evaluation. Both methods follow the configuration in Appendix~\ref{app:protocol}. \method{} completes training in 8 h 2 min 21 s, compared with 8 h 21 min 3 s for \directopd{}, reducing total training time by 18 min 42 s (3.73\%). The corresponding mean step times are 289.41 s and 300.63 s, a reduction of 11.22 s per step. Total GPU-hours decrease from 66.81 to 64.31.

% Generated by scripts/build_result_tables.py; edit data/efficiency_results.json.
\begin{table}[!htbp]
  \caption{\textbf{Training-time comparison on Qwen3-8B math.} Total training time over 100 training steps on eight H20 GPUs, excluding evaluation.}
  \label{tab:efficiency}
  \centering
  \begingroup
  \small
  \setlength{\tabcolsep}{4pt}
  \renewcommand{\arraystretch}{1.12}
  \begin{tabularx}{\linewidth}{@{}l>{\raggedleft\arraybackslash}X>{\raggedleft\arraybackslash}X>{\raggedleft\arraybackslash}X@{}}
    \toprule
    Method & Time (h:m:s) & Seconds/step & GPU-hours \\
    \midrule
    \directopd{} & 8:21:03 & 300.63 & 66.81 \\
    \textbf{Waypoint-OPD} & 8:02:21 & 289.41 & 64.31 \\
    \bottomrule
  \end{tabularx}
  \par\vspace{3pt}
  \parbox{\linewidth}{\footnotesize GPU-hours are computed as eight times the reported training duration in hours and exclude evaluation.}
  \endgroup
\end{table}

Relative to \directopd{}, \method{} adds one scoring pass with the frozen Waypoint checkpoint while using the same number of student rollouts. With this additional source of temporal consistency information, \method{} achieves a lower measured training time in this setting.

\begin{figure}[!htbp]
  \centering
  \begin{minipage}[t]{0.49\linewidth}
    \centering
    \includegraphics[width=\linewidth]{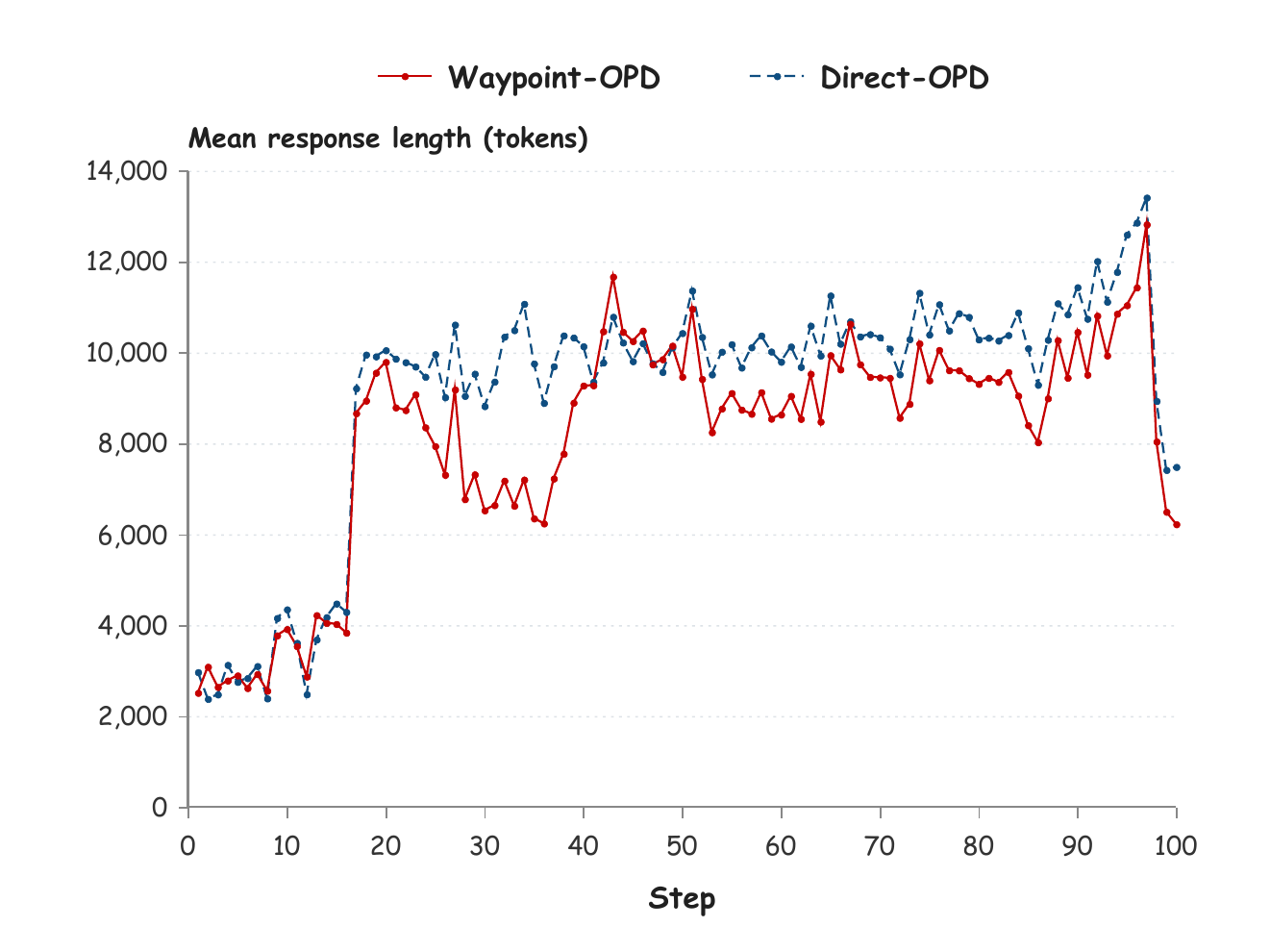}
    \par\smallskip
    {\small (a) Mean response length}
  \end{minipage}\hfill
  \begin{minipage}[t]{0.49\linewidth}
    \centering
    \includegraphics[width=\linewidth]{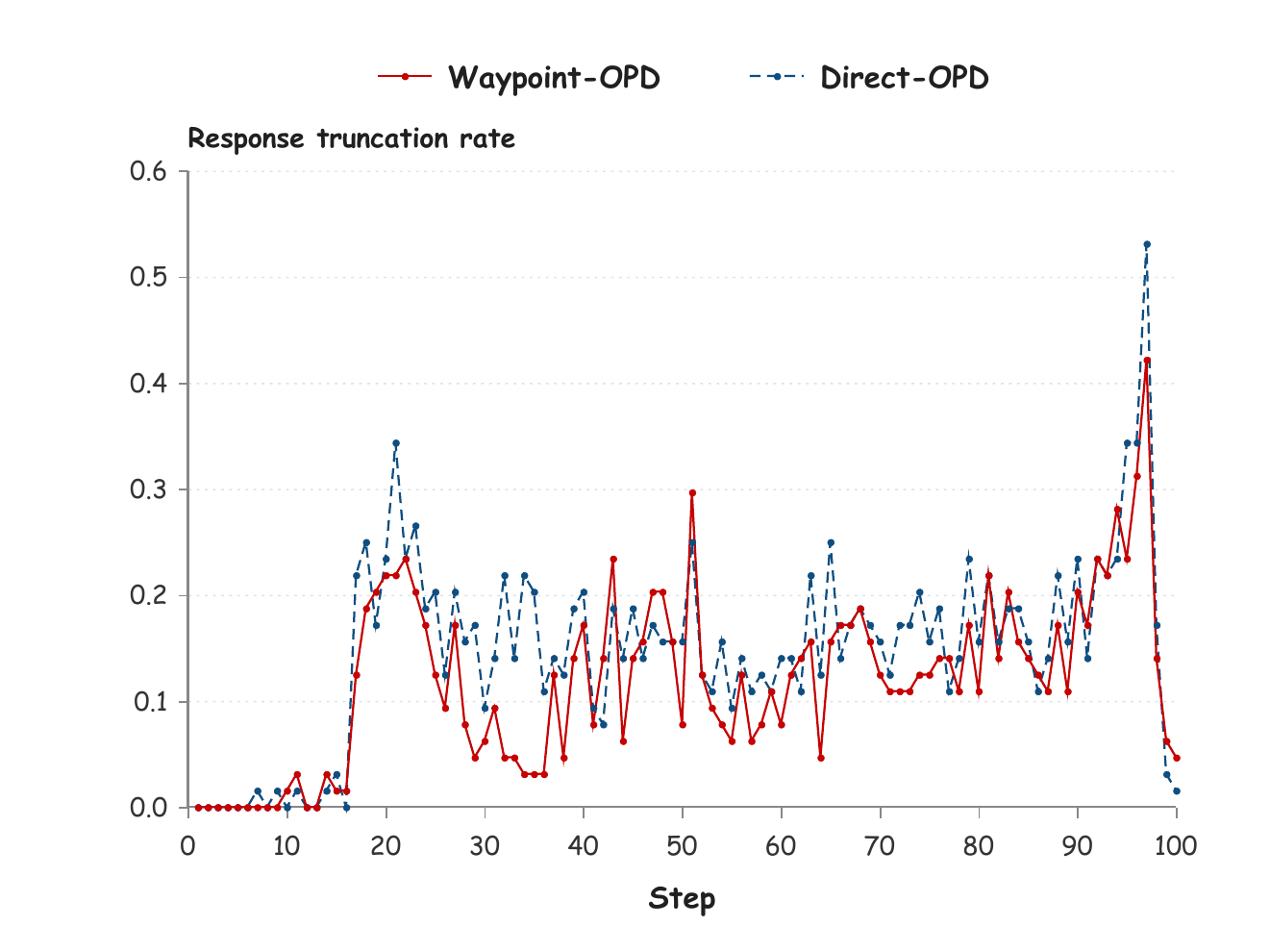}
    \par\smallskip
    {\small (b) Response truncation rate}
  \end{minipage}
  \caption{\textbf{Response length and response truncation rate.} The two panels compare Qwen3-8B trained with \method{} (solid red) and \directopd{} (dashed blue) over 100 training steps. (a) Mean number of tokens per rollout response
  at each training step. (b) Fraction of rollout responses reaching the maximum length of 16,384 tokens; 0.2 corresponds to 20\%.}
  \label{fig:response-length}
\end{figure}

Figure~\ref{fig:response-length} compares rollout generation over the same 100 training steps. Response length is the mean number of tokens per rollout response at each training step, and response truncation rate is the fraction of rollout responses reaching the maximum length of 16,384 tokens. \method{} has a lower mean response length at 87 of the 100 steps and a lower response truncation rate at 62 steps, with equal rates at 18 steps. Averaged uniformly across the 100 steps, the response truncation rate decreases from 14.83\% for \directopd{} to 11.97\% for \method{}, a reduction of 2.86 percentage points. These results show that \method{} produces shorter responses at most training steps and reaches the response-length limit less frequently on average.

\end{document}